\documentclass[letterpaper, 10 pt, conference]{ieeeconf}  

\IEEEoverridecommandlockouts
\usepackage{cite}
\usepackage{amsmath,amssymb,amsfonts}
\usepackage{algorithmic}
\usepackage{graphicx}
\usepackage{textcomp}
\usepackage{xcolor}
\usepackage{eqexpl}
\usepackage{autobreak}
\usepackage{comment}

\usepackage{stfloats} 
\usepackage{amsmath}

\usepackage[utf8]{inputenc}
\usepackage[ruled,longend]{algorithm2e}
\usepackage{textgreek}
\usepackage{amssymb}
\usepackage{float}
\usepackage{svg}
\usepackage{tikz}
\usepackage{pgfplots}
\usepackage{booktabs}
\usepackage{authblk}
\usepackage{float}
\usepackage{mathtools}

\graphicspath{{Figures/}}
\pgfplotsset{compat=1.17}

\newlength\imagewidth
\newlength\imagescale

\newcommand{\ra}[1]{\renewcommand{\arraystretch}{#1}}

\def\BibTeX{{\rm B\kern-.05em{\sc i\kern-.025em b}\kern-.08em
    T\kern-.1667em\lower.7ex\hbox{E}\kern-.125emX}}

\begin{document}

\title{Prediction Based Decision Making for Autonomous Highway Driving\\
}

\author{Mustafa Yildirim$^{1}$ Sajjad Mozaffari $^{2}$ Luc McCutcheon$^{1}$  Mehrdad Dianati $^{2}$ Alireza Tamaddoni-Nezhad $^{3}$ \\ Saber Fallah$^{1}$
\thanks{$^{1}$Mustafa Yildirim, Luc McCutcheon and Saber Fallah are with CAV-Lab, Department of Mechanical Engineering Sciences, University of Surrey,
        {\tt\small \{m.yildirim,lm01065, s.fallah\}@surrey.ac.uk}}%
\thanks{$^{2}$Alireza Tamaddoni-Nezhad is with Department of Computer Science, University of Surrey 
        {\tt\small a.tamaddoni-nezhad@surrey.ac.uk}}%
\thanks{$^{3}$Sajjad Mozaffari and Mehrdad Dianati are with Warwick Manufacturing Group, University of Warwick
        {\tt\small \{sajjad.mozaffari,M.Dianati\}@warwick.ac.uk}}%
}

\maketitle

\begin{abstract}

Autonomous driving decision-making is a challenging task due to the inherent complexity and uncertainty in traffic. For example, adjacent vehicles may change their lane or overtake at any time to pass a slow vehicle or to help traffic flow. Anticipating the intention of surrounding vehicles, estimating their future states and integrating them into the decision-making process of an automated vehicle can enhance the reliability of autonomous driving in complex driving scenarios. This paper proposes a Prediction-based Deep Reinforcement Learning (PDRL) decision-making model that considers the manoeuvre intentions of surrounding vehicles in the decision-making process for highway driving. The model is trained using real traffic data and tested in various traffic conditions through a simulation platform. The results show that the proposed PDRL model improves the decision-making performance compared to a Deep Reinforcement Learning (DRL) model by decreasing collision numbers, resulting in safer driving.

\end{abstract}


\section{Introduction}

Highly Automated Vehicles (HAVs) are a rapidly developing technology with a vital role in society. The aim of technology is to improve driving safety for drivers, passengers and pedestrians, reduce traffic congestion and decrease fuel consumption.
One major functionality of HAVs is Driving Decision-Making (DDM) which is responsible for making decisions on when to take action(s), what action(s) to perform and how to perform an action (s). These decisions can be one at a time, such as to reach a specific velocity or sequential such as completing an overtake manoeuvre (accelerate-lane change) \cite{ross2014robot}. HAVs need to cooperate with other traffic participants to make safe and reliable decisions in continuously varying traffic conditions.

Rule-based methods are frequently used for decision-making, specifically for lane-change manoeuvres \cite{niehaus1994probability}. Automated vehicles change lanes based on designed rules such as the gap acceptance model \cite{gipps1986model} or the minimise-overall-braking induced by lane changes (MOBIL) model \cite{kesting2007general}. Nevertheless, vehicles using rule-based methods are conservative in making decisions in a variety of traffic conditions, which may adversely impact traffic flow \cite{Li2019a}. Furthermore, the rules are inflexible and cannot handle unseen driving situations. Methods of DDM should be adaptable and generalisable to cope with uncertainties and unseen driving situations.

More advanced decision-making techniques compared to rule-based techniques have been studied for highway driving. Decision trees\cite{quinlan1986induction} and random forest\cite{ho1995random} methods are hugely depend on data and prone to over-fit. The support vector machine (SVM)\cite{liu2019novel} is sensitive to noise, and a minute change in data might lead to a different result. The game theory approach \cite{meng2016dynamic} and Fuzzy-logic\cite{balal2016binary} are applicable when traffic levels are low, but the complexity of the problem increases proportionally to the number of vehicles. Monte Carlo Tree Search (MCTS) \cite{silver2010monte} is one promising technique for the decision-making problem of highway driving\cite{gonzalez2019human}, but a limited number of search branches is considered in most studies due to the computational cost. 
Combining MCTS with a neural network, known as deep MCTS, helps to better guide the sampling towards the most relevant sub-trees \cite{Hoel2020} and improves the computational efficiency. Despite their promising results, most of the aforementioned techniques suffer from lack of generalisation or adaptability in unseen driving conditions.

Recently, Reinforcement Learning (RL) has received significant attention from researchers as a powerful technique to solve complex, uncertain DDM problems due to its strong adaptability and generalisation. At first, value-based techniques such as Q-learning\cite {Watkins1992q} were used by researchers for DDM due to its simplicity \cite{li2015reinforcement}. However, the performance of such methods for complex and high dimensional driving situations such as dense traffic is limited because of the curse of dimensionality. The combination of Q-learning with neural networks addresses this problem and paves the path for using Q-learning for complex DDM applications\cite{mukadam2017tactical}. Early implementations of Deep Q-Networks (DQN) was used for lane-keeping control of a vehicle in a race track \cite{sallab2017deep,Wolf2017LearningHT}. Later, DQN algorithms were used by different research works to make decisions in highway driving scenarios \cite{8951131}. For instance,  using a quantile regression DQN (QR-DQN) \cite{pmlr-v80-dabney18a}, Min et al. introduced a sensor fusion structure controller to decide lane-keeping, lane changing, and acceleration control in their study \cite{min2018deep}. Mo et al. studied \cite{Mo2019a} the challenging scenario of oncoming traffic by implementing Double DQN \cite{van2016deep} as a DDM to perform the lane-change manoeuvre, but decision-making is supported by rule-based Time to Collision (TTC) \cite{hyden1996traffic} rewards which calculates collision time based on constant speed. 
Shi et al. \cite{Shi2019} proposed a hierarchical decision-making model, where a DQN model decides when to perform a lane change manoeuvre, and another Q function approximator decides how to complete the manoeuvre. Mirchevska et al. \cite{Mirchevska2018}, and Shu et. al\cite{mohammadhasani2021reinforcement} implemented a safety check to DQN outputs for the execution of action to prevent the collision. Safety check considers output based on safety distance to prevent collision if in case after lane change, the lead vehicle performs sudden brake and cause ego vehicle to collide. The real-time performance of the aforementioned DDM techniques suffers from the lack of foreseeing the other vehicles' intentions. It is known that driving intentions of surrounding vehicles significantly influence the decisions made by HAVs and have to be included in the DDM process.

In separate works, researchers contributed to predicting the intentions of surrounding vehicles. A study by Gindele et al. deals with simplified traffic models and makes accurate predictions based on the Partially Observable Markov Decision Processes (POMDP) \cite{Gindele2015} . Alizadeh et al.\cite{8917192} studied the same problem by integrating noise when measuring other vehicles' positions and compared the result with the rule-based lane change model Mobil \cite{Treiber2013}. In another study \cite{wei2014behavioral}, authors predicted surrounding cars velocity and used this information for path planning. Based on this prediction, they decreased the calculation cost by 90\%.  Jiang et al. \cite{Jiang2019} considered the estimation of other vehicles' intentions for trajectory planning in their work, but the prediction only considered whether the target lane vehicle is cooperative or aggressive when performing the lane change manoeuvre. Kochenderfer et. al. anticipated driver cooperativeness for merging scenario \cite{bouton2019cooperation}. 
Recent studies\cite{bahram2014prediction,ulbrich2015towards}, integrated the intention prediction module to decision-making process to improve the performance. For instance, Gonzalez et al. proposed a human-like decision making \cite{gonzalez2019human} by implementing a belief tree to estimate lane change of other drivers using MCTS, however the sampling time of tree search and the lack of tractability was the limitation of MCTS. In another study\cite{schildbach2015scenario}, target vehicle trajectory is predicted based on three actions such as we have used in our model, but the prediction horizon considers only two seconds ahead just for the target vehicle on the defined lane, whereas our study contains five seconds prediction horizon for surrounding six vehicles. The studies in literature either predict the behaviour as a general (such as aggressive or passive) or generate a belief tree based on initial prediction/assumption. In contrast to the literature mentioned above, our study continuously considers lane change prediction and updates the prediction for each observation state for a longer time zone as long as the surrounding vehicles are within the radar range.

This paper focuses on how to integrate intention predictions of other vehicles into the DDM process of a HAV to make safer decisions through DQN methods. The paper proposes a prediction-based DDM method for lane-change highway driving scenarios by integrating the intention of surrounding vehicles as a time to lane change in the decision-making model, which helps decreasing potential collisions. In this paper, different DQN techniques are used to formulate the DDM problem and their performance's are compared with and without a prediction module.  

The paper is structured as follows. 
In section 2, the proposed decision-making methodology is explained. 
Section 3 clarifies network parameters as well as training and test evaluations. Section 4 discusses and evaluates the results, and section 5 concludes the study.

\section{Proposed Method}
\subsection{Assumptions}

\begin{figure}[h]
\centerline{\includegraphics[scale=0.42]{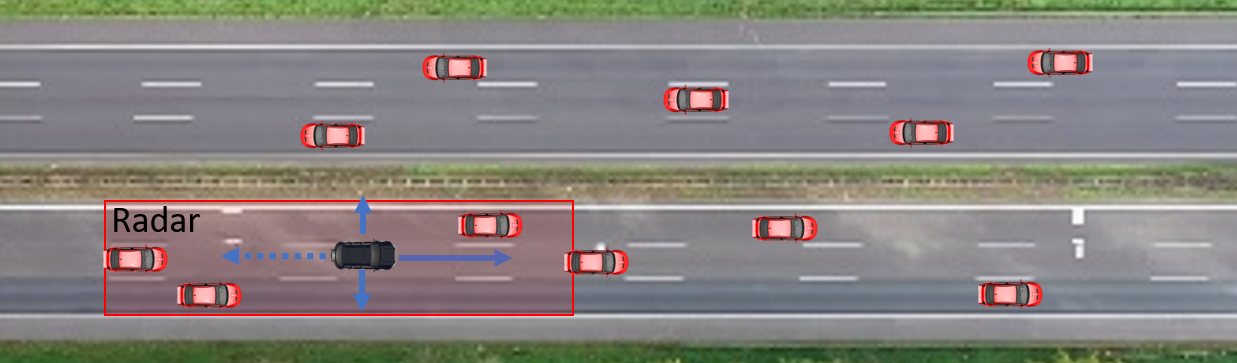}}
\caption{Ego vehicle driving on highway }
\label{fig:control}
\end{figure}

We assume that an Ego vehicle (EV) and target vehicles (TVs) are driving on a three-lane straight highway. The EV is equipped with a 360-degree radar sensor that can detect the position and velocity of surrounding vehicles within $R=250$ m distance to the EV. Using a bird-eye view (BEV) camera, we can observe the environment for the prediction of up to six adjacent vehicles as shown in Figure \ref{fig:control}.
The Ego vehicle utilizes all this information to predict surrounding cars' intention (whether they are going to perform a lane change or stay on the current lane).

\begin{figure*}[h]
\centerline{\includegraphics[scale=0.378]{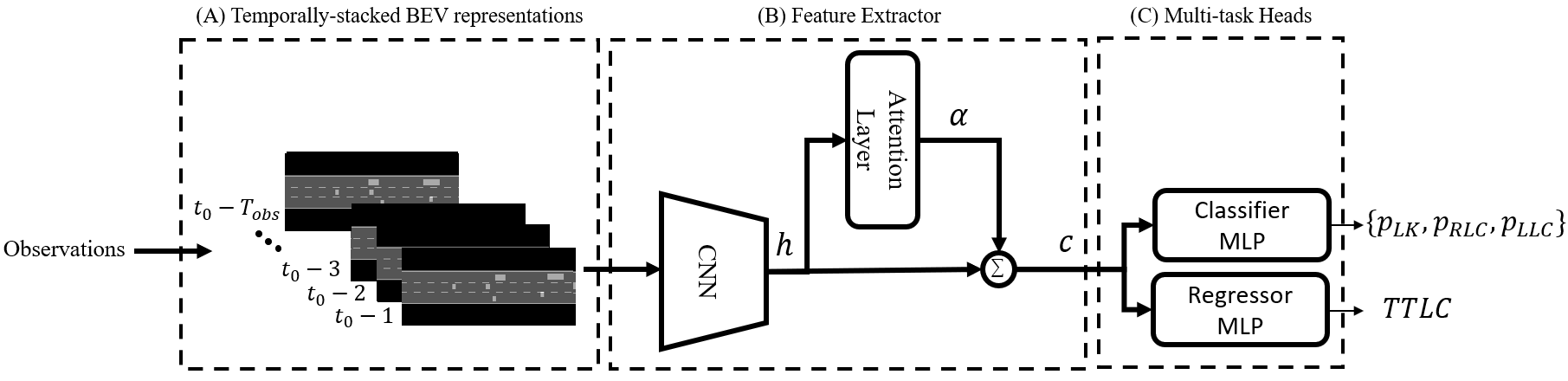}}
\caption{An overview of the prediction model \cite{9740533}}
\label{prediction_model}
\end{figure*}

The manoeuvre intention predictor consists of two parts as shown in Figure \ref{prediction_model}.
The first part is a classifier that identifies the intention of the other cars as ${P_{LK}, P_{RLC}, P_{LLC}}$  where LK represents a lane-keeping manoeuvre, RLC and LLC represent lane change manoeuvres to the right and to the left, respectively. The second part is a regressor which estimates Time-to-Lane Change (TTLC), t $\epsilon$ T as $\{5,4,3,2,1,0\}$ where numbers represent time in seconds when classifier actions occur.

By adding this information, the decision-making model has distance, velocity and prediction information of nearby vehicles. Having all this information, the decision-making model controls the EV lane change decision such as left lane change, right lane change or lane-keeping and reacts instantly in vehicle dynamics limit by maintaining  driving safety.

\subsection{Decision-Making Model using Reinforcement Learning }
A Markov Decision Process (MDP) is the mathematical framework for decision making that is the basis of the RL problem formulation. \cite{Bellman1977a}. An MDP is composed of an action set A, a state set S, a reward function R and a transition model P(s'$|$s, a). An RL agent learns to maximise expected cumulative reward in an MDP by taking an action a $\epsilon$ A, which reaches a new state s' $\epsilon$ S and receives a reward r. The agent updates its policy $\pi$ to maximise future cumulative rewards and the process is repeated until the agent has sufficiently optimised the policy to achieve the highest score in the environment.

Cumulative rewards is the summation of a sequence of rewards received. A discount factor $\gamma$ $\epsilon$ [0,1] is then applied at each time-step to trade off the importance of immediate rewards over long term rewards.

The state-space includes the position and velocity of EV, surrounding vehicles' distance to EV,  their velocities, and intentions. The EV's action space is defined as a left lane change, right lane change or lane-keeping  A=\{LLC, RLC, LK\}. The goal of RL is to find optimal policy $\pi$ that maximizes total future rewards:

\begin{equation}
R(\pi, r)=\mathbb{E}_{\pi}\left[\sum_{t=0} \gamma^{t} r\left(s_{t}, a_{t}\right)\right]
\end{equation}
When solving sequential decision problems, estimations are learned for the optimal value of each action, which is defined as the expected value of the rewards in the future when taking that action and following the optimal policy accordingly. The policy is used to predict the value for each action in A for the current state. This action-value function is formulated as:

\begin{equation}
    Q_{\pi}(s,a)=\mathbb{E}[R_{1} + \gamma R_{2} + ... | S_0 = s, A_0 = a, \pi]
\end{equation}

Actions can then be chosen greedily with respect to this action-value function, or alternatively, exploratory actions can be taken given the current observation s $\epsilon$ S.

A neural network is then trained by minimising the loss function below at each iteration ${i}$, optimising the network weights $\theta$,

\begin{equation}
\begin{aligned}
L_{i}\left(\theta_{i}\right)=\mathbb{E}_{s, a \sim \rho(\cdot)}\left[\left(y_{i}-Q\left(s, a ; \theta_{i}\right)\right)^{2}\right]\\
y_{i}=\mathbb{E}_{s^{\prime} \sim \varepsilon}\left[r+\gamma \max _{a^{\prime}} Q\left(s^{\prime}, a^{\prime} ; \theta_{i-1}\right) \mid s, a\right]
\label{eqn:Function Aproximation Loss}
\end{aligned}
\end{equation}

$y_i$ is the target and $\rho(s,a)$ is the probability distribution of states and actions. The differential of the loss function with respect to the weights gives us the following gradient.

\begin{multline}
\nabla_{\theta_{i}} L_{i}\left(\theta_{i}\right)=\mathbb{E}_{s, a \sim p(\cdot) ; s^{\prime} \sim \mathcal{E}} 
\bigg[\left(r+\gamma \max _{a^{\prime}} Q\left(s^{\prime}, a^{\prime} ; \theta_{i-1}\right)\right) \\-
Q\left(s, a ; \theta_{i}\right) \nabla_{\theta_{i}} Q\left(s, a ; \theta_{i}\right)\bigg]
\label{eqn:Loss Gradient}
\end{multline}

This gradient is followed through Adam (Adaptive Moment Estimation) gradient decent on each iteration.

\subsection{Reward Functions}
\begin{itemize}
    \item Collision Penalty: Collision is an undesirable situation and to prevent this event, the highest penalty is given for collision status.
    \item End of Track: If the agent completes the track and reaches the goal, it receives a positive reward value.
    \item Lane Change Penalty: To prevent an unnecessary lane change, the agent receives a small penalty if there is a lane change, since the value is very low comparing to other rewards and penalties, it does not effect the goal of the agent but prevents it from taking unnecessary actions.

\end{itemize}

Combining all reward functions defines the total reward function as follows:

\begin{multline}
{Reward\: Function} =  R_{End\:of\:Track} - R_{Lane\:Change}\\
      \qquad - R_{Collision}
\label{eqn:Reward_Function}
\end{multline}

\begin{figure*}[h]
\centerline{\includegraphics[scale=0.55]{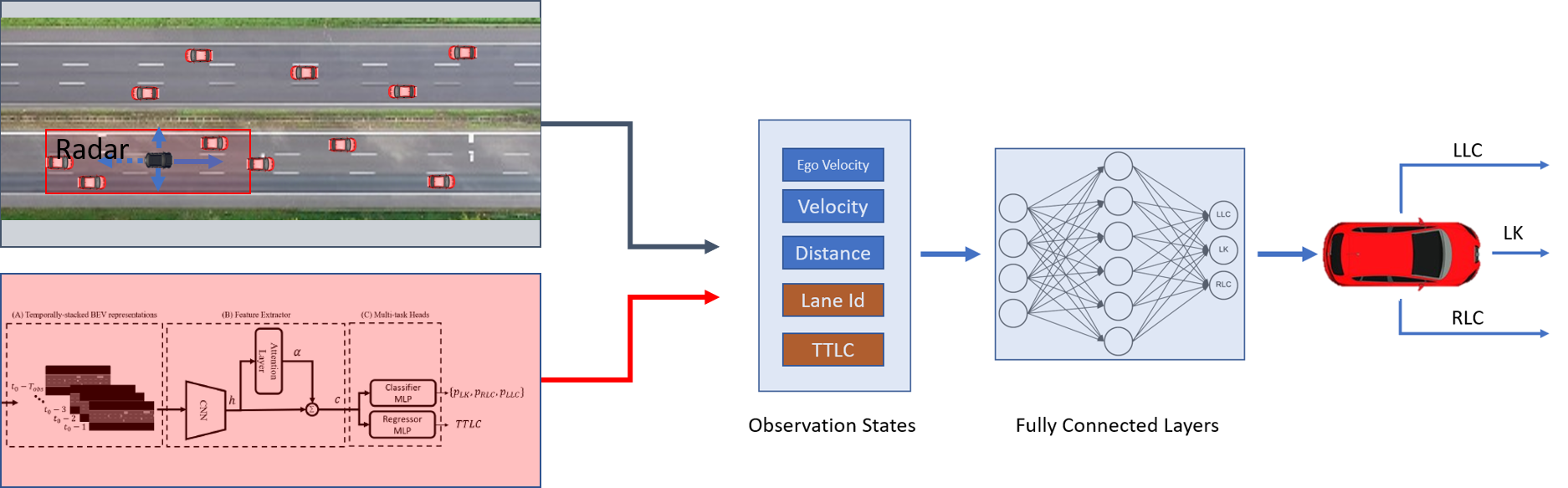}}
\caption{Proposed TTLC  method network structure}
\label{fig:proposed}
\end{figure*}

\subsection{Intention Prediction Model}

The proposed decision-making model (Figure \ref{fig:proposed}) is combined with predicted actions of surrounding cars. The information for the lane change manoeuvre fed into the fully connected layer for each state, in additional to velocity and distance information.
The intention of other cars to lane change is implemented as Time-To-Lane-Change (TTLC) which is introduced in the study of Mozaffari et al.\cite{9740533}. Figure~\ref{prediction_model} shows an overview of the prediction model and a summary of key processing steps of the prediction model is provided below:
\begin{itemize}
\item \textbf{BEV input representation}: First, the states of the target vehicle and its surrounding vehicles at each time-step are used to generate a simplified top-down view representation of the driving environment. This representation includes the vehicle bounding boxes, the road marking and the drivable area, each encoded with a specified value. A temporal channel-wise stack of BEV representation for the past few time-steps creates the input data at the current time-step.
\item \textbf{Attention-based Feature Extractor}: A six-layer Convolutional Neural Network (CNN) \cite{kalchbrenner2014convolutional} is used to extract relevant features for the prediction task from the stacked BEV representations. To enhance the feature extraction performance, a special attention mechanism is designed to selectively focus on one of the quarter areas around the target vehicle, namely, front-right, front-left, back-right and back-left. The introduction of the attention mechanism helps the prediction model to identify the informative area of the surrounding environment in each prediction query.
\item \textbf{Multi-task Heads}: A multi-task learning approach has been used in dealing with lane change prediction problem. The lane change prediction problem is defined as estimating the likelihood of lane change manoeuvres (i.e., right lane change, left lane change and lane-keeping) within a prediction horizon and regressing the Time-To-Lane-Change. A separate multilayer perceptron head is considered for the likelihood estimation and regression parts, although both heads use the extracted features by the attention-based CNN model. The overall model is trained using a weighted sum of a mean squared loss for the regression part and cross-entropy loss for the classifier (i.e., likelihood estimation).

\end{itemize}

\subsection{Vehicle Longitudinal and Lateral Controllers}
The EV lateral controls are defined as discrete actions and the decision making model perform actions to change or keep lane. The Intelligent Driver Model (IDM) \cite{Treiber2000} controls the longitudinal movement.

\subsubsection{Actions}

The agent has three actions: left lane change, right lane change, and stay at the current lane. These actions are discrete, and the decision-making model performs the best action based on observed inputs. The IDM controls the EV's acceleration and velocity.

\subsubsection{IDM}
The IDM controls vehicle acceleration and deceleration based on the distance and velocity between the EV and the vehicle in front of the EV, and adjust vehicle velocity accordingly. The IDM automatically prevents collision by observing distance and decreasing velocity if the EV is too close to the front vehicle. In our model, acceleration is calculated based on the below equations and parameters \cite{naghshvar2018risk} shown in the Table \ref{tab:IDM} for the IDM model.

\begin{equation}
a=a_{\max }\left[1-\bigg(\frac{v}{v_{\text {desired }}}\bigg)^{4}-\left(\frac{s^{*}\left(v, v_{\text {lead }}\right)}{s}\right)^{2}\right]
\end{equation}

\begin{multline}
  s^{*}\left(v, v_{\text {lead}}\right)=\max \bigg( s_{0}, v \rho+\frac{1}{2} a_{\max}\rho^{2}+ \\
  + \frac{\left(v+\rho a_{\max }\right)^{2}}{2 b_{\text {safe }}}- \frac{\left(v_{\text{lead }}\right)^{2}}{2 b_{\max }} \bigg)
\end{multline}

\begin{table}[ht]\centering
\caption{ IDM Parameters}
\ra{1.3}
\begin{tabular}{@{}lccc@{}}\toprule
 
$Parameter$  & $Value$  \\ \midrule
Minimum Distance ($s_{0}$) & 5 m  \\
Desired Velocity ($V_{desired}$) & 130 km/h  \\
Maximum Acceleration ($a_{max}$) & 3  m/s  \\
Maximum Deceleration ($b_{max}$) & 5 m/s \\
Safe Deceleration ($b_{safe}$) & 4 m/s \\
Response Time ($\rho$) & 0.25 s \\
\bottomrule
\end{tabular}
\label{tab:IDM}
\end{table}

\subsection{Reinforcement learning Methods for DDM}

The Deep Q-Networks (DQN) differs from Q-Learning by using deep networks composed of layers and neurons instead of a Q-table. State, action, reward, and next state are the main elements of DQN. Based on observation states, the agent performs random actions using an epsilon greedy policy to explore the environment. Based on these states and actions, the agent reach a new state and obtains a reward. The action-value (Q-value) is calculated based on rewards obtained using the Bellman equation. The targets are the Q-values of each of the actions and the input would be the state that the agent is in and the intention prediction of surrounding cars. This is an iterative process where the agent stores learning information (state, action, reward, next state, terminal state flag) in the replay buffer. The agent then learns to optimise the Q-function to maximise future expected cumulative reward using a random batch of stored transitions from the replay buffer. Repeating this process helps the agent to maximize its cumulative reward, allowing the agent to perform the best action for a new state. The algorithm for Deep Q-learning with experience replay is shown below. Due to the variety of methods, only the base algorithm \cite{Mnih2015b} is given here.

\begin{algorithm}

Initialize replay memory $\mathcal{D}$

Initialize action-value function ${Q}$ with random weights

\For{ \rm{$episode = 1$, } ${M}$} {
        Initialise sequence $s_1 = \{x_1\}$ and $\phi_1 = \phi(s_1)$
        
        \For{ $t = 1$, ${T}$} {
        Select random action $a_t$ with probability $\epsilon$\\
        select $a_t = max_a Q*(\phi(s_t),a;\theta)$ based on $1-\epsilon$\\
        Execute action $a_t$ and obtain reward $r_t$ and state $x_{t+1}$
        Set $s_{t+1} = s_t,a_t,x_{t+1}$ and preprocess $\phi_{t+1} = \phi(s_{t+1})$\\
        Store transition $(\phi_t,a_t,r_t,\phi_{t+1})$  in  $\mathcal{D}$\\
        Sample random minibatch of transitions $(\phi_j,a_j,r_j,\phi_{j+1})$ from $\mathcal{D}$
        \[
        Set y_j = \begin{cases}
        r_j :\text{for terminal } \phi_{j+1}\\
        r_j + \gamma max_a' Q(\phi_{j+1}, a';\theta)\\ \hspace{1em}:\text{for non-terminal } \phi_{j + 1}
        \end{cases}\]
        Perform gradient decent step on $(y_i - Q(\phi_j,a_j;\theta))^2$ according to equation \ref{eqn:Loss Gradient}

        } 
      
}

\caption{DQN with Experience Replay}
\end{algorithm}

\begin{multline}
  Q(s_t,a_t)\;\leftarrow\;Q(s_t,a_t) +\\
  +\alpha [ r_{t+1} + \gamma \max_a Q(s_{t+1},a) - Q(s_t,a_t) ]
\end{multline}

The DQN method, by its nature,  might have tendency to overestimate  Q-values. To overcome this drawback, many improvements have been proposed in the literature. To extend our study and compare the methods proposed in literature, five different DQN variants have been implemented. The variants included in our study are:

\begin{figure*}[htbp]
\centerline{\includegraphics[scale=0.5]{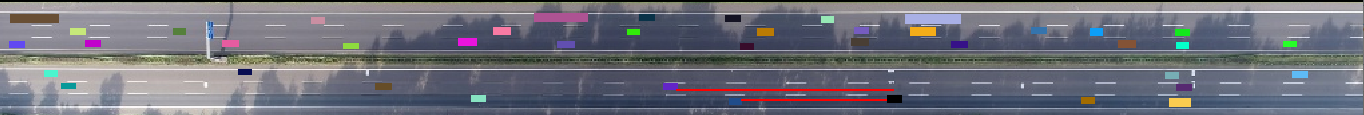}}
\caption{Simulation of HighD traffic}
\label{fig:highd}
\end{figure*}

\begin{itemize}
\item DQN \cite{Mnih2015b}: Have tendency to overestimate Q-values.
\item Double DQN\cite{van2016deep}: Uses the target network to calculate the Q-value to solve the overestimation problem.
\item Averaged DQN\cite{Ansehel2017a}:Provides more stable training and reduces approximation errors by taking an average of the last five values.
\item Duelling DQN\cite{Wang2016}: Separates networks as two layers such as advantage and value.
\item Noisy Network\cite{fortunato2017noisy}: Adds noise to weights to improve exploration efficiency.


\end{itemize}

\section{Performance Evaluation}

There are many open-source platforms to simulate traffic environments, such as Carla \cite{dosovitskiy2017carla}, AirSim \cite{shah2018airsim} and Sumo\cite{lopez2018microscopic} but generating simulated traffic based on real traffic data was not straightforward to implement to these platforms. Since we aimed to simulate an agent on a real-world application by predicting real drivers intentions, these platforms were also unsuitable. Therefore as shown in Figure \ref{fig:highd}, we have generated a Pygame \cite{mcgugan2007beginning} based traffic simulation environment.

\subsection{Dataset}

Highway driving has been studied in various works based on two main datasets NGSim \cite{yeo2008oversaturated} and HighD \cite{krajewski2018highd}. Both datasets contain all necessary information such as lateral and longitudinal position, velocity and acceleration information for each vehicle in traffic. The HighD dataset is collected using a drone capturing a 420 m long part of the German highway. A wide-angle camera observes the position of vehicles and provides accurate data in occluded traffic. In this study, training and testing were performed based on the HighD dataset. Fifteen different track's data were used for training, and a further 15 were used for testing from the 60 track dataset.

Since this traffic is not rule-based or randomly-generated, it reflects real tests results. On the other hand, using datasets has some drawbacks, for example other vehicles can not sense the EV and for that reason, there are inevitable collisions caused by other vehicles.

\begin{figure}[h]
    \centering
    \pgfmathsetlength{\imagewidth}{\linewidth}%
    \pgfmathsetlength{\imagescale}{\imagewidth/524}%
    \begin{tikzpicture}[x=\imagescale,y=-\imagescale]
        \node[anchor=north west] at (0,0) {\includegraphics[width=\imagewidth]{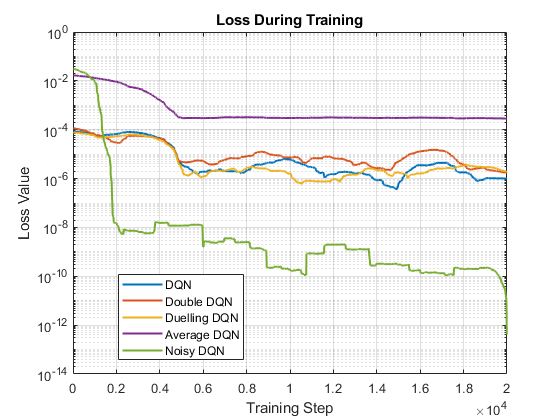}};
    \end{tikzpicture}
    \caption{Loss comparison of methods}
    \label{fig:loss_all}
\end{figure}

\begin{figure}[h]
    \centering
    \pgfmathsetlength{\imagewidth}{\linewidth}%
    \pgfmathsetlength{\imagescale}{\imagewidth/524}%
    \begin{tikzpicture}[x=\imagescale,y=-\imagescale]
        \node[anchor=north west] at (0,0) {\includegraphics[width=\imagewidth]{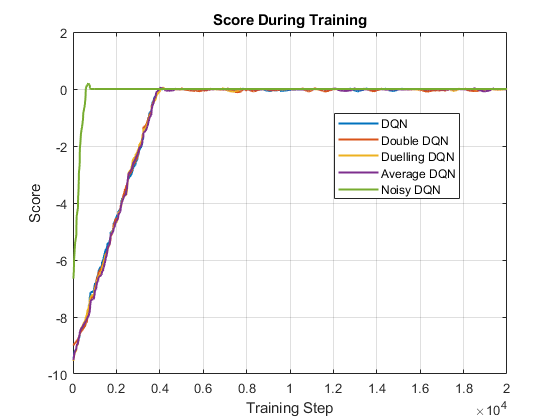}};
    \end{tikzpicture}
    \caption{Score comparison of methods}
    \label{fig:score_all}
\end{figure}


\subsection{Network \& Hyperparameters}
Our initial evaluations were performed to determine hyperparameters for optimal and fast convergence. Parameters were chosen based on initial runs (see Table \ref{tab1}). In addition to these hyperparameters, analysis was performed to use the dropout feature \cite{srivastava2014dropout}. Decision making model with the dropout could not be converged to the stable loss value.  This feature was also tested as an experiment but it did not contribute to the decision-making model and therefore dropout was excluded.

The network was composed of 3 fully connected layers; the first two layers include 128 nodes, and the final layer is formed of 3 nodes to represent agent action space.
The Pytorch library was used for neural networks computations. Networks were optimized using the ADAM algorithm, which is computationally efficient and converges rapidly \cite{kingma2014adam}. The discount rate was set to $\gamma = 0.95$, and the learning rate to $\alpha = 10e-5$. The size of the experience replay memory is 10 000. The batch size for stochastic gradient descent is 32. Epsilon-greedy exploration policy was used  for all methods except Noisy networks, starting from $\epsilon = 1$  and decreasing to $\epsilon = 0.01$ as a minimum value.

\begin{table}[ht]\centering
\caption{ Model Hyperparameters}
\ra{1.3}
\begin{tabular}{@{}lccc@{}}\toprule
 
$Hyperparameter$  & $Value$  \\ \midrule
Learning Rate ($\alpha$) & 10e-5  \\
Discount Factor ($\gamma$) & 0.95  \\
Epsilon min ($\epsilon$) & 0.01  \\
Memory & 10e4 \\
Batch Size & 32  \\
\bottomrule
\end{tabular}
\label{tab1}
\end{table}

\begin{table*}[ht]\centering
\caption{ Collision number comparison for all 5 DQN extension for base and proposed TTLC method by using Ground Truth and prediction model}
\ra{1.3}
\begin{tabular}{@{}rcccccccc@{}}\toprule
& \multicolumn{1}{c}{Base} & \phantom{abc}& \multicolumn{2}{c}{Ground Truth TTLC} &
\phantom{abc} & \multicolumn{3}{c}{Predicted TTLC}\\
\cmidrule{2-2} \cmidrule{4-5} \cmidrule{7-9}
& $Collision Number$  && $Collision Number$ & $Improvement$  \% &  && $Collision Number$ & $Improvement$ \% \\ \midrule
DQN & 12.73 & & 10.84 & 17.43 &&&  12.44   & 2.33 \\
Double DQN & 13.11 & & 12.6  & 4.04 &&& 12.02   &  9.06 \\
Averaged DQN & 12.33 & & 9.53 & 29.3 &&& 9.93  & 19.46 \\
Duelling DQN & 13.82 & & 11.4 & 21.22 &&& 12.57    & 9.94 \\
Noisy DQN & 11.55 & & 10.93 & 5.6 &&& 11.53   &  0.17 \\

\bottomrule
\end{tabular}\\
\label{tab2}
\end{table*}

\subsection{Experiments}
Initially, 20 000 episodes were chosen since methods were converging between 12 000 and 14 000 episodes, but after hyperparameter optimizations, methods converged around 7 000. Training time differs between methods. Base DQN training finished in 10 h, Noisy DQN  and Double DQN were completed in 11 h, Duelling DQN in 16 h and Average DQN in 19 h. TTLC methods added an additional two hours for each method.

The loss graph of methods in Figure \ref{fig:loss_all} shows that the most smooth curve belongs to Average DQN; although it has a higher loss value than other methods, it is the most stable. On the other hand, Noisy DQN has the lowest loss value and has more fluctuations than other methods. In general, DQN, Double DQN and Duelling DQN have similar values in between. DQN is slightly lower than others, and Double DQN is noticeably higher than Duel DQN. Comparison of methods in terms of the score shows that all methods have reached the maximum score and have a stable policy as seen in Figure \ref{fig:score_all}. It is observed that Noisy DQN has reached the maximum score sooner than other DQN variants, and it shows that Noisy DQN is more sample efficient than others. This is because it explores the environment differently than the epsilon-greedy approach.

The tests were performed 45 times, and the average collision number is measured for comparison. Both the base method and proposed method have been tested on the same tracks. As it can be seen from Table \ref{tab2}, the proposed TTLC approach improved the performance of decision-making for highway driving comparing to base methods. Comparing all, among the base methods, Noisy DQN showed the best performance as having the lowest collision number on average as 11.55, followed by Averaged DQN as 12.33 and DQN as 12.73. On the other hand, the proposed method contributed most to the Averaged DQN as 29.2\% improvement, and collision number is decreased to 9.53 and followed by DQN with 10.84 collisions on average and then Noisy DQN as 10.93 collisions. The proposed TTLC method has contributed a significant improvement to the results. The most noticeable contribution was to the Averaged DQN, followed by Duelling DQN as 21.22\% and DQN as 17.43\%. On average, for the five methods proposed, there was 15.51\% less collisions in general with ground truth data. It can be inferred from Table \ref{tab2} that the predicted TTLC shows less contribution for each method comparing to the ground truth. The predicted TTLC decreased the collision rate by 8.19\%  when averaged over all methods.

\section{Discussion}
The results of this study show that the best performance is obtained by  Noisy DQN for base approaches and Averaged DQN for the proposed method. The novel improvements and extensions were implemented to DQN, such as Double and Duelling DQN, but these two extensions could not perform better than DQN. A general assumption about these extensions is; Double DQN and Duelling DQN methods have better performance than DQN, as shown in \cite{Wang2016} and \cite{VanHasselt2016}. However, this is not consistent in all environments, as seen in \cite{Wang2016}. Reinforcement learning approaches distinctly depend on the environment and action space of the agent since everything is built on the interaction of these two. According to \cite{Wang2016}, Duel DQN showed -100\% worse performance than DQN on the Freeway environment in which an agent attempts to avoid other traffic participants and pass through the highway. In this environment, similar to our case, the agent has only three actions and avoids collisions. As a result, Duelling DQN is not always better than DQN. In limited action space, it shows reduced performance compared to DQN. The advantage of the duelling network lies in its ability to approximate values efficiently. When the number of actions is high, this advantage over single-stream Q networks increases \cite{tan2018cs234}. It shows superior performance if there is a possibility of the agent having multiple actions per time step, such as the Atlantis environment \cite{bellemare2013arcade}.

The proposed TTLC method improved the result significantly and provided a safer autonomous drive. The contribution of TTLC varied in different DQN extensions and Averaged TTLC showed the best improvement among the TTLC methods. The prediction horizon overlaps with the Average DQN since the Average DQN takes an average of the last five values and the prediction model considers ongoing five seconds for each time step.

The different methods cause variations in the neural network weight updates and give rise to the fluctuations between methods. In addition, a limitation of our model is that the radar can only detect up to 6 cars at a time which causes uncertainty and impedes the decision-making model for the cases where more than six surrounding vehicles are present as they are not visible to the EV. These two factors could be the reason for the variance between the DQN extensions. However, despite differences between the extensions, on average, the proposed TTLC contributed to having 15.51\% less collision than the base methods. This improvement clearly shows our proposed TTLC methods' contribution compared to the base methods and therefore concludes that TTLC promises a better approach for real-world applications such as highway driving.

\section{Conclusion}
In this work, we have tested various DQN methods for autonomous highway driving. A decision-making model integrated with a prediction model has been proposed to improve highway driving safety. The intention of surrounding cars is integrated into the decision-making model and compared with base methods. The proposed method shows that predicting surrounding cars' intention to lane change decreases collision possibility and provides safer driving than base approaches. This study has revealed that Averaged DQN TTLC showed the best performance within the methods in this environment. Another outcome of the study is that the action space influences the contribution of DQN variants to the performance.

\bibliographystyle{unsrt}
\bibliography{library}

\end{document}